\documentclass[letterpaper]{article} 
\usepackage{aaai23}  
\usepackage{times}  
\usepackage{helvet}  
\usepackage{courier}  
\usepackage[hyphens]{url}  
\usepackage{graphicx} 
\usepackage{natbib}  
\usepackage{caption} 
\usepackage{amsmath, amsthm, amssymb, amsfonts}
\usepackage{booktabs}
\usepackage{lipsum}
\usepackage{xcolor}

\title{Semi-Automated Construction of Food Composition Knowledge Base}
\author {
    Jason Youn,\equalcontrib\textsuperscript{\rm 1,2,3}
    Fangzhou Li,\equalcontrib\textsuperscript{\rm 1,2,3}
    Ilias Tagkopoulos\textsuperscript{\rm 1,2,3}
}
\affiliations {
    \textsuperscript{\rm 1} Department of Computer Science, University of California, Davis, CA 95616, USA.\\
    \textsuperscript{\rm 2}Genome Center, University of California, Davis, CA 95616, USA.\\
    \textsuperscript{\rm 3}USDA/NSF AI Institute for Next Generation Food Systems (AIFS), University of California, Davis, CA 95616, USA.\\
    {jyoun, fzli, itagkopoulos}@ucdavis.edu
}

\begin{document}

\maketitle

\begin{abstract}

A food composition knowledge base, which stores the essential phyto-, micro-, and macro-nutrients of foods is useful for both research and industrial applications. Although many existing knowledge bases attempt to curate such information, they are often limited by time-consuming manual curation processes. Outside of the food science domain, natural language processing methods that utilize pre-trained language models have recently shown promising results for extracting knowledge from unstructured text. In this work, we propose a semi-automated framework for constructing a knowledge base of food composition from the scientific literature available online. To this end, we utilize a pre-trained BioBERT language model in an active learning setup that allows the optimal use of limited training data. Our work demonstrates how human-in-the-loop models are a step toward AI-assisted food systems that scale well to the ever-increasing big data.

\end{abstract}

\section{Introduction}

Constructing high-quality knowledge bases of foods is crucial for various needs like understanding the impact of dietary intake on human health and enabling personalized diet recommendations \cite{Jacobs2007,Young2003}. There exist multiple knowledge bases that attempt to curate food composition information such as FoodData Central (406,999 foods and $\sim$300 key nutrients) \cite{usda2019}, FooDB (797 foods and 15,750 chemicals) \cite{tmic2017}, KNApSAcK (24,704 plant species and 62,647 metabolites) \cite{shinbo2006knapsack}, and Phenol-Explorer (459 foods and 501 polyphenols) \cite{rothwell2013phenol}. In an effort to curate the relationship between chemicals and human health, knowledge bases like CTD \cite{Davis2022} and KEGG \cite{kanehisa2000kegg} curate information about the interactions among chemicals, genes, and/or disease entities, while other resources, including ChemFont \cite{wishart2022chemfont} and GO \cite{ashburner2000gene}, are dedicated to creating an ontology of chemicals. Yet, existing approaches to creating and expanding these knowledge bases are often bottlenecked by the need for time-consuming manual annotation processes that often require the expertise of domain experts.

\begin{figure}[t]
\centering
\includegraphics[width=0.95\columnwidth]{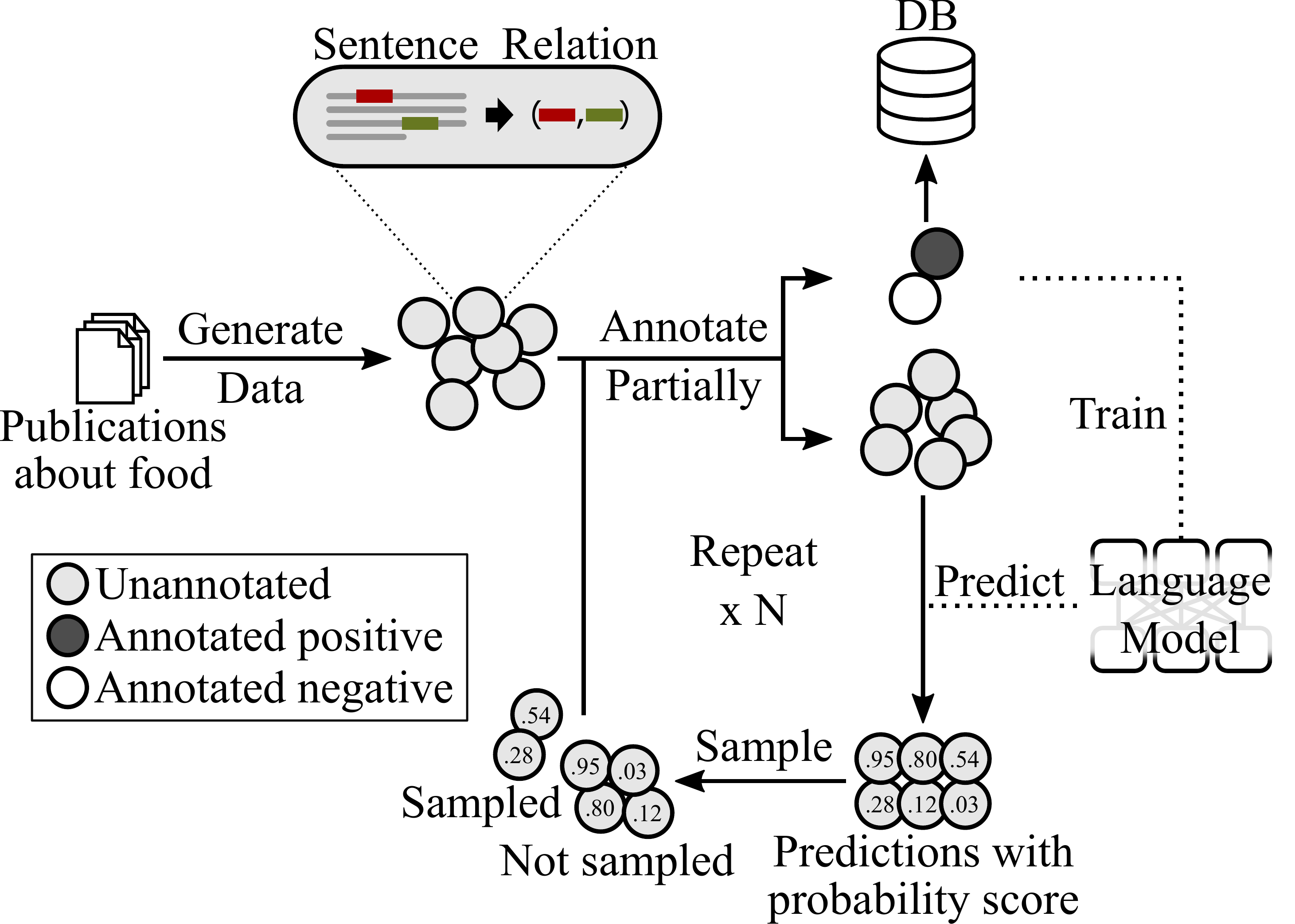}
\caption{Overarching pipeline for semi-automated construction of a food knowledge base from online publications using language models with active learning. From the sentences in the literature that mention both foods (red) and chemicals (green), we extract relations where food and chemical entities are connected by the \textit{contains} relation. We then use active learning with a language model to partially annotate and train over $N=10$ rounds.}
\label{fig:main}
\end{figure}

While knowledge bases have traditionally been curated manually from text data \cite{kotova2019automated}, recent approaches utilize deep learning-based state-of-the-art relation extraction (RE) models for constructing knowledge bases \cite{jiang2020complex}. RE is a task in natural language processing (NLP) that extracts semantic relations between entities in natural language sentences \cite{bach2007review} (e.g., given a sentence `Joe Biden is the president of United States’, an RE model extracts a relation of `isPresidentOf’ between the entities `Joe Biden’ and `United States’). However, these deep learning-based approaches often require many labeled training data \cite{lecun2015deep}, which is often not feasible in fields like food science where the data annotation procedure is expensive. To address such issue, active learning methods, which use the model to choose the data that can most efficiently improve its performance and therefore reduce the amount of data needed for achieving the desired training outcome \cite{settles2009active}, have been widely used in various fields such as natural language processing \cite{shen2004multi,longpre2022active,rotman2022multi}, computer vision \cite{Coleman2022,chen2022active,Yu2021}, and studies in biology \cite{Wang2020}. Furthermore, recent works suggest that semi-automatic models are desirable for tasks related to knowledge bases as machines, though faster than human annotators, often yield low recall \cite{wang2022human,zhuang2017hike}.

In this work, we propose a semi-automated framework for constructing a knowledge base of food composition information using active learning of language models (Figure \ref{fig:main}). Training and evaluating 100 runs of the proposed active learning strategy each with a unique random seed shows that language models are able to extract correct relations from the sentence with high confidence (precision $=0.92\pm0.04$, recall $=0.82\pm0.07$, and F1 $=0.87\pm0.04$). We also found that using the proposed active learning sampling strategy accelerates new knowledge discovery by $21.0\%\pm0.05\%$. All code and instructions on how to reproduce the results can be found in \url{https://github.com/ibpa/SemiAutomatedFoodKBC}.

\section{Proposed Method}

\textbf{Data Generation.} We downloaded 1,226 food names (both commonly used and scientific) from FooDB \cite{tmic2017} and used them to query LitSense \cite{allot2019litsense}, a sentence-level search system provided by National Center for Biotechnology Information (NCBI) for biomedical literature from PubMed and PubMed Central (PMC). We used the search query template \textit{`food\_name contains'} as we empirically found that the returned sentences had the most food-chemical relations. In addition to the food and chemical entities already tagged and returned by LitSense for each sentence, we manually generated a list of 70 common food parts and used a rule-based named entity recognition (NER) approach to strictly match the food part entities. Note that since LitSense returns all species in the NCBI taxonomy even if they are not food, we dropped non-food species by keeping only the species in the 1,226 FooDB food names. Therefore, for each sentence $s \in S$ returned by LitSense, three sets of entities $F$, $P$, and $C$ for foods, food parts, and chemicals, respectively, were recognized. Finally, we extracted a set of relations $R$ for $s$ as

\begin{equation}
\begin{gathered}
R = \{template(f, p, c) \mid \forall (f,p,c) \in F \times P \times C \} \\
\cup \{template(f, c) \mid \forall (f, c) \in F \times C \},
\label{equ:relations}
\end{gathered}
\end{equation}

where $template( \cdot )$ is a function that takes as input entities, with or without food part entity, and outputs a \textit{contains} relation between entities. For example, $template$(apple, vitamin A) = \textit{apple contains vitamin A} and $template$(apple, skin, vitamin A) = \textit{apple skin contains vitamin A}, if there is a food part entity. This process resulted in 85,839 sentence-relation (SR) pairs with 21,313 unique sentences.

\textbf{Data Annotation.} We randomly selected train, validation, and test datasets from these SR pairs, while making sure there were no duplicate sentences or relations across the datasets that could cause bias during training and evaluating (Table \ref{tab:dataset}). During the manual annotation process, two annotators were asked to assign three possible classes \textit{positive}, \textit{negative}, and \textit{skip}. The \textit{positive} class was assigned when there was enough evidence in the sentence to support that the paired relation was true. Otherwise, the \textit{negative} class was assigned. We assigned the class \textit{skip} if the NER by LitSense was not performed correctly, where we eventually discarded the skipped SR pairs. To ensure that the annotation quality was high, we kept only the \textit{positive} and \textit{negative} SR pairs whose annotation results were consensus between the two annotators.

\begin{table}[t]
 \caption{Statistics of the dataset used for training and evaluation. Note that the training data of 1,000 sentence-relation(SR) pairs are distributed evenly across N=10 rounds of active learning, while the validation and test set is held-out for consistency.}
  \centering
  \begin{tabular}{cccc}
    \toprule
     & train & val & test \\
    \midrule
    \# of SR pairs & 1,000 & 300 & 300 \\
    \# of positive SR pairs & 453 & 116 & 129 \\
    \# of negative SR pairs & 547 & 184 & 171 \\
    \toprule
    \# of unique sentences & 747 & 174 & 157 \\
    \# of unique relations & 1,000 & 300 & 300 \\
    \toprule
    \# of entities & 537 & 175 & 169 \\
    \# of food entities & 288 & 95 & 86 \\
    \# of chemical entities & 249 & 80 & 83 \\
    \bottomrule
  \end{tabular}
  \label{tab:dataset}
\end{table}

\textbf{Language Model.} We used the BioBERT \cite{lee2020biobert} language model that is based on the original BERT \cite{devlin2018bert} model but trained using biomedical domain corpora PubMed and PMC, sharing a similar domain of the text returned by LitSense \cite{allot2019litsense}. We formatted input to the model by concatenating the sentence and relation strings separated by the [SEP] token. We fine-tuned BioBERT using the binary classification scheme, with the grid search of 8 possible hyperparameter combinations (learning rate = $\{2 \times 10\textsuperscript{-5}, 5 \times 10\textsuperscript{-5}\}$, batch size = $\{16, 32\}$, and epochs = $\{3, 4\}$) using the validation set. Finally, the hyperparameter set with the highest precision score was selected for the active learning step that we will discuss in the next section.

\begin{table*}[ht]
\caption{Performance metrics for 100 runs of active learning (uncertainty sampling) compared to the random sampling.}
\centering
\setlength{\tabcolsep}{3.5pt}
\begin{tabular}{ccccccccccc}
    \toprule\multicolumn{1}{c}{} & \multicolumn{10}{c}{Rounds} \\
    \cmidrule(r){2-11}
    Metric & 1 & 2 & 3 & 4 & 5 & 6 & 7 & 8 & 9 & 10 \\
    \midrule
    \multicolumn{11}{l}{\emph{Active Learning (Uncertain)}} \\
    \midrule
    Precision & .87$\pm$.09 & .86$\pm$.05 & .87$\pm$.05 & .88$\pm$.05 & .89$\pm$.05 & .89$\pm$.04 & .91$\pm$.04 & .91$\pm$.04 & .92$\pm$.04 & .92$\pm$.04 \\
    Recall & .51$\pm$.29 & .72$\pm$.13 & .76$\pm$.10 & .78$\pm$.10 & .76$\pm$.12 & .78$\pm$.09 & .81$\pm$.07 & .80$\pm$.08 & .81$\pm$.08 & .82$\pm$.07 \\
    F1 & .57$\pm$.28 & .77$\pm$.08 & .81$\pm$.07 & .82$\pm$.01 & .81$\pm$.07 & .83$\pm$.06 & .85$\pm$.05 & .85$\pm$.05 & .86$\pm$.04 & .87$\pm$.04 \\
    Accuracy & .75$\pm$.10 & .83$\pm$.04 & .85$\pm$.04 & .86$\pm$.04 & .85$\pm$.04 & .86$\pm$.04 & .88$\pm$.03 & .88$\pm$.03 & .89$\pm$.03 & .89$\pm$.03 \\
    Specificity & .93$\pm$.07 & .91$\pm$.04 & .91$\pm$.04 & .92$\pm$.04 & .93$\pm$.04 & .93$\pm$.03 & .94$\pm$.03 & .94$\pm$.03 & .95$\pm$.03 & .94$\pm$.03 \\
    \midrule
    \multicolumn{11}{l}{\emph{Random}} \\
    \midrule
    Precision & .88$\pm$.07 & .88$\pm$.05 & .88$\pm$.04 & .88$\pm$.04 & .90$\pm$.04 & .90$\pm$.04 & .91$\pm$.04 & .91$\pm$.04 & .92$\pm$.04 & .92$\pm$.04 \\
    Recall & .49$\pm$.28 & .70$\pm$.17 & .76$\pm$.10 & .78$\pm$.11 & .79$\pm$.09 & .79$\pm$.08 & .81$\pm$.07 & .82$\pm$.08 & .81$\pm$.07 & .82$\pm$.06 \\
    F1 & .56$\pm$.27 & .77$\pm$.12 & .81$\pm$.06 & .82$\pm$.06 & .84$\pm$.05 & .84$\pm$.04 & .86$\pm$.04 & .86$\pm$.04 & .86$\pm$.04 & .87$\pm$.04 \\
    Accuracy & .74$\pm$.10 & .83$\pm$.06 & .85$\pm$.03 & .86$\pm$.03 & .87$\pm$.03 & .87$\pm$.03 & .88$\pm$.03 & .89$\pm$.03 & .89$\pm$.03 & .89$\pm$.03 \\
    Specificity & .93$\pm$.06 & .92$\pm$.05 & .92$\pm$.04 & .92$\pm$.04 & .93$\pm$.03 & .93$\pm$.03 & .94$\pm$.03 & .94$\pm$.03 & .95$\pm$.03 & .95$\pm$.03 \\
    \bottomrule
\end{tabular}
\label{tab:performance}
\end{table*}

\textbf{Active Learning.} We used the pool-based active learning strategy as defined by Settles \cite{settles2009active}, where we split the total training pool of 1,000 annotated SR pairs (Table \ref{tab:dataset}) equally into 10 active learning rounds. For the first round, we randomly sampled 100 training SR pairs from the pool. We trained the language models using these 100 SR pairs and then predicted the probability of being a \textit{positive} class, denoted as $p$, for the remaining 900 SR pairs. From round 2, we selected the data using the \textit{uncertainty} sampling scheme, where we sampled the SR pairs closest to the model decision boundary. The uncertainty score for each SR pair was calculated as $min(1-p,p)$, and then the \textit{uncertainty} scheme would sample 100 SR pairs with the highest uncertainty scores. These newly sampled data were added to the previous training data and used to train the next rounds. We repeated this process until round 10 when all training data were sampled and consumed by the model. We refer to this whole process as a single run of active learning, and we repeated this active learning run 100 times to provide statistical significance for our results.

\begin{figure}[t]
\centering
\includegraphics[width=1.0\columnwidth]{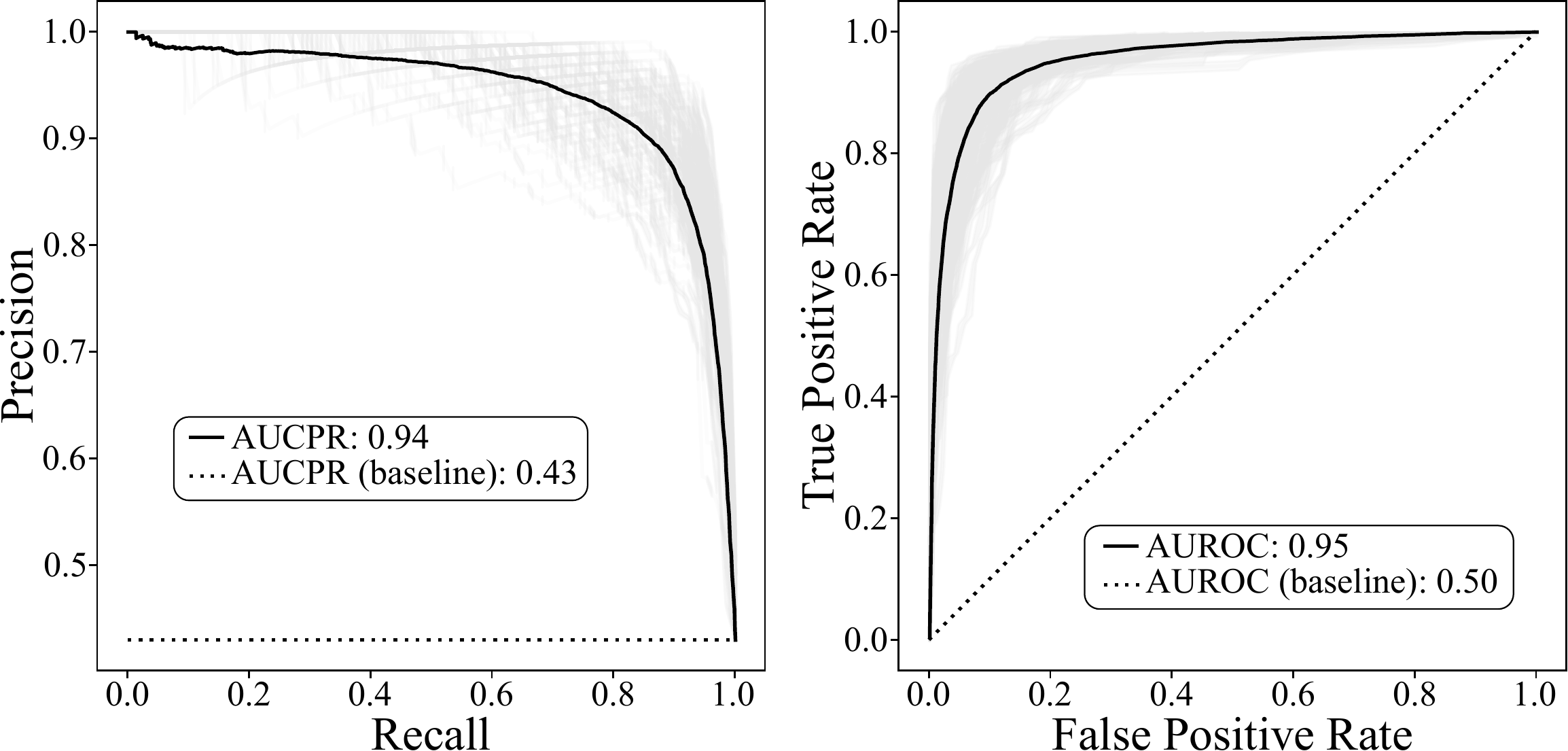}
\caption{Precision-recall (left) and receiver operating characteristic (right) curves of the models at round 10 with active learning. Gray lines indicate the curves for 100 runs, and the black lines are plotted by concatenating the model probabilities of the test data from these 100 runs. The baseline refers to a classifier that always predicts the minority class (positive).}
\label{fig:pr_roc}
\end{figure}

\section{Experiments and Results}

\textbf{Settings.} We used 2 $\times$ Nvidia RTX A5000 GPUs for the training and inference of the BioBERT language models, one GPU for each of the active learning strategies, (i.e., uncertainty sampling and random sampling.) The entire procedure (100 runs $\times$ 10 rounds for each active learning strategy) took approximately 90 hours, where around 85\% of the time was used for model training and the rest for data sampling and evaluation. The model pipeline was implemented with PyTorch \cite{paszke2019pytorch} and HuggingFace's transformer library \cite{wolf2019huggingface}.

\textbf{Active Learning Results.} The performance metrics obtained from 100 different runs of the active learning each with 10 rounds and different random seeds are shown in Figure \ref{tab:performance}. Compared to the first round, precision at the final 10th round increased by 6.2\% ($0.87\pm0.09$ vs. $0.92\pm0.04$, respectively), recall by 62.0\% ($0.51\pm0.29$ vs. $0.82\pm0.07$, respectively), and F1 by 50.9\% ($0.57\pm0.28$ vs. $0.87\pm0.04$, respectively). However, compared to the random sampling active learning strategy which chooses the samples to train on for the subsequent round randomly (gray line and box plots in Figure \ref{tab:performance}), we did not observe any statistical difference between the performance metrics (\textit{p}-value $> 0.05$) except for precision at round 2 ($0.88\pm0.05$ vs. $0.86\pm0.05$, respectively, \textit{p}-value $= 3.4\times10\textsuperscript{-2}$), recall at round 5 ($0.79\pm0.09$ vs. $0.76\pm0.12$, respectively, \textit{p}-value $= 2.9\times10\textsuperscript{-2}$), and F1 at round 5 ($0.84\pm0.05$ vs. $0.81\pm0.07$, respectively, \textit{p}-value $= 6.8\times10\textsuperscript{-3}$). The performance metrics start to show an insignificant difference in their average values compared to the final round at round 8 for precision ($0.91\pm0.04$ vs. $0.92\pm0.04$, respectively, \textit{p}-value $= 1.6\times10\textsuperscript{-1}$), round 7 for recall ($0.81\pm0.07$ vs. $0.82\pm0.07$, respectively, \textit{p}-value $= 2.5\times10\textsuperscript{-1}$), and round 9 for F1($0.86\pm0.04$ vs. $0.87\pm0.04$, respectively, \textit{p}-value $= 3.2\times10\textsuperscript{-1}$). Note that all \textit{p}-values were calculated using the two-sided t-test. The models at the final round that was trained using the complete training data have AUCPR $= 0.94$ and AUROC $= 0.95$ as shown in Figure \ref{fig:pr_roc}.

\begin{figure}[t]
\centering
\includegraphics[width=0.7\columnwidth]{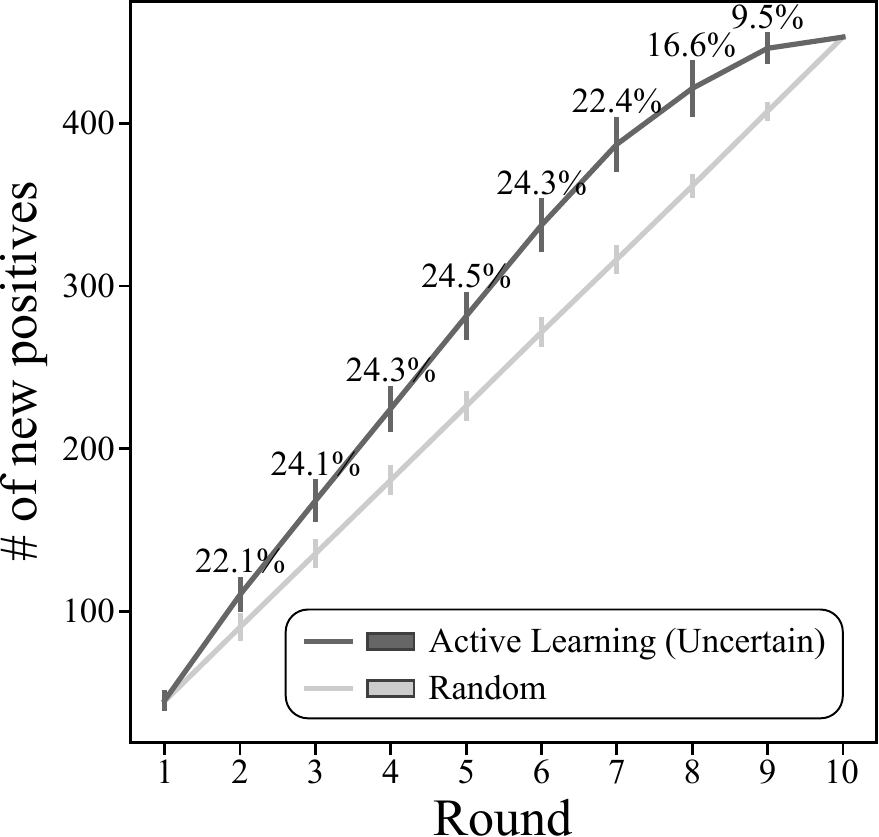}
\caption{Rate of discovery of new knowledge within the training data compared between the uncertainty active learning sampling strategy and the random sampling strategy. Percentage value denotes the increase of relative abundance of positive data for each round from random to uncertainty active learning method.}
\label{fig:num_positives}
\end{figure}

Although both the random sampling strategy and the uncertainty active learning strategy discover the same set of 453 positives in the training pool with 1,000 SR pairs (Table \ref{tab:dataset}) at the final round, the uncertainty active learning strategy discovers positive data $21.0\%\pm0.05\%$ faster on average during the intermediate rounds (rounds 2 through 9) compared to the random sampling strategy that discovers new knowledge in each round of training linearly (Figure \ref{fig:num_positives}).

\section{Conclusion}

In this work, we presented a semi-automated framework for constructing a food composition knowledge base from scientific literature using pre-trained language models supported by active learning. Although the active learning sampling strategy that selects the uncertain samples around the probability of 0.5 has not shown statistically significant predictive performance improvement over the random sampling approach, we found that the uncertainty sampling approach was able to find new positive data faster, therefore leading to the creation of knowledge base in an accelerated manner than the random sampling. In future work, we plan to test additional sampling strategies like disproportionate stratified sampling and one that only samples data with the highest probability scores. We also plan to train the model with a bigger dataset, create a knowledge graph of food-chemical information enriched with ontological relationships like taxonomy and chemical classification, and perform link prediction \cite{yao2019kg,youn2022kglm} on the knowledge graph to discover novel food-chemical relations.

\bibliography{fa_references}

\section{Acknowledgments}

We would like to thank the members of the Tagkopoulos lab for their suggestions, and Gabriel Simmons for the initial discussions. This work was supported by the USDA-NIFA AI Institute for Next Generation Food Systems (AIFS), USDA-NIFA award number 2020-67021-32855, and the NIEHS grant P42ES004699 to I.T. All computational analyses were performed and the figures were generated by J.Y. and F.L., and J.Y., F.L., and I.T. contributed to the critical analysis and wrote the paper. I.T. supervised all aspects of the project.

\end{document}